\documentclass{article}

\pdfoutput=1
\usepackage{fullpage}
\usepackage{amssymb}
\usepackage[utf8]{inputenc}
\usepackage{amsmath}
\usepackage{amsfonts}
\usepackage{alltt}
\usepackage{tabularx,colortbl}
\usepackage{multirow}
\usepackage{color}
\usepackage[pdftex]{hyperref}
\usepackage{algorithm}
\usepackage{algorithmic}
\usepackage{graphicx}
\usepackage{subfigure}
\usepackage{listings}
\usepackage[numbers,sort&compress]{natbib}
\usepackage[english]{babel}
\usepackage{authblk}

\newcommand{\argmin}{\operatornamewithlimits{arg\ min}}

\definecolor{Orange}{rgb}{1,0.64,0}
\definecolor{lgray}{rgb}{0.85,0.85,0.85}

\begin{document}

\title{Building pattern recognition applications with the {SPARE} library}

\author[1]{Lorenzo Livi\thanks{llivi@scs.ryerson.ca}\thanks{Corresponding author}}
\author[2]{Guido Del Vescovo\thanks{delvescovo@uniroma1.it}}
\author[2]{Antonello Rizzi\thanks{antonello.rizzi@uniroma1.it}}
\author[2]{Fabio Massimo {Frattale Mascioli}\thanks{mascioli@infocom.uniroma1.it}}
\affil[1]{Dept. of Computer Science, Ryerson University, 350 Victoria Street, Toronto, ON M5B 2K3, Canada}
\affil[2]{Dept. of Information Engineering, Electronics, and Telecommunications, SAPIENZA University of Rome, Via Eudossiana 18, 00184 Rome, Italy}
\renewcommand\Authands{, and }
\providecommand{\keywords}[1]{\textbf{\textit{Index terms---}} #1}

\maketitle

\begin{abstract}
This paper presents the SPARE C++ library, an open source software tool conceived to build pattern recognition and soft computing systems. The library follows the requirement of the generality: most of the implemented algorithms are able to process user-defined input data types transparently, such as labeled graphs and sequences of objects, as well as standard numeric vectors.
Here we present a high-level picture of the SPARE library characteristics, focusing instead on the specific practical possibility of constructing pattern recognition systems for different input data types. In particular, as a proof of concept, we discuss two application instances involving clustering of real-valued multidimensional sequences and classification of labeled graphs.\\
\end{abstract}

\section{Introduction}

The classic pattern recognition discipline is essentially conceived for the $\mathbb{R}^{n}$ vector space geometry. However, many interesting applications, coming for instance from computational biology, multimedia intelligent processing, and computer vision, deal with structured patterns, such as images, audio and video sequences, strings of characters, and labeled graphs \cite{delvescovo07OnlineHandwritingRecognitionbySymbolicHistogramsApproach,tun+dhar+palumbo+giuliani2006,giuliani2002}. Usually, in order to take advantage of the existing data-driven modeling systems, each pattern of a structured domain $\mathcal{X}$ is reduced to a set of real-valued features by adopting a preprocessing function, tailored on the specific application at hand \cite{livi+delvescovo+rizzi_seriation+gradis,delvescovo_rlgradis_2012,odse,gralg_2012,seriation+gradis_lncs_2012,riesen+bunke2010,Queiroz20133383}.
The design of this preprocessing procedure is a critical task, since useful information could be lost due to an excessive information compression.
For this reason, it could be useful to design pattern recognition systems able to deal directly with structured or unconventional input domains. Consequently, the availability of effective models and algorithms able to represent and process a set of samples belonging to $\mathcal{X}$ is fundamental \cite{izakian2015fuzzy,jiang+mnger+bunke2001,ferrer10Generalizedmediangraphcomputationbymeansgraphembeddingvectorspaces,bardaji+ferrer+sanfeliu2010}.
When dealing with cluster analysis, the abstraction of a cluster of objects by a representative model is useful for two main reasons. The first reason is about computational issues. For example, having to compare a particular pattern with the representative instead of each single element in the cluster often results in a significative speed up. 
The second reason deals with the ability of a learning system to create abstract knowledge, a fundamental feature of the cerebral cortex in biological brains. In fact, the human ability to synthesize concepts from a set of objects plays a key role in human reasoning and thus in any machine learning procedure.

SPARE -- Something for PAttern REcognition -- is a C++ library conceived as an essential software framework for rapid development of applications of pattern recognition and computational intelligence.
SPARE is based on the \textit{concept-driven} template meta-programming technique \cite{abrahams05C++templatemetaprogrammingconceptstoolstechniquesboostbeyond}, aiming to provide basic classes and procedures for rapid application development when dealing with clustering algorithms, inductive modeling systems (classification and function approximation), and optimization problems by neural networks, fuzzy logic, and evolutionary computation.
The set of concepts and classes has been thought in order to capture the essence of some basic yet important algorithms, and reflects the fact that even the most complex data-driven modeling system can be conceived as an organized ensemble of small code blocks, each of them implementing specific functionalities with a standard interface. Designing a library with concepts in mind permits to create a set of classes that are not relegated to a fixed hierarchy: concepts are \textit{non-restricting} design choices, which in essence relaxes the development process.
The meta-algorithms implemented in SPARE can be easily used to build pattern recognition applications dealing with customizable (i.e., user-defined) data types, without any strict necessity to represent objects as feature-based patterns. Moreover, the template-based approach simplifies the interoperation with other state-of-the-art software libraries \cite{dlib09,Chang:2011:LLS:1961189.1961199} and with the code written by the user.
Being written in standard C++, SPARE code is also efficient and fairly portable to several major operating systems and architectures. SPARE provides multi-thread implementations of different algorithms. This is an important features of modern hardware, which should be exploited in any computationally intensive application.

SPARE is an open source software library, released under the GNU GPL 3.0 license. The main site of the project is located at \href{https://sourceforge.net/p/libspare/home/Spare/}{https://sourceforge.net/p/libspare/home/Spare/}, where it is possible to obtain an up-to-date version of the SPARE library, as well as all the available documentation, tutorials, and some focused benchmarks considering other well-known software libraries.
We decided to release SPARE as an open source project since, in our opinion, using open source software in research projects allows to easily reproduce and verify experimental results, therefore improving the transparency of the research itself.
Moreover, it allows a faster exchange of knowledge between different research groups.

In this paper, we present the high-level design and functionalities of the SPARE library together with two practical application instances.
The first one concerns the clustering of a set of sequences of multidimensional real-valued vectors, and aims to analyze the modeling capability of one of the implemented cluster representative techniques in SPARE. The second one discusses how to develop and test a graph-based classification system with SPARE. A benchmark of four graph matching algorithms on a batch of 15 synthetic classification problem instances for labeled graphs is contextually analyzed.
The herein reported applications are introduced only as a proof of concept to show the versatility of the SPARE library in designing pattern recognition systems for different data types. Therefore, the presented experimental results should not be considered as ``research results''.
Technical details regarding the library are omitted in this paper, referring the reader to the SPARE documentation.
The involved theoretical and methodological aspects are not discussed as well; we provide literature references for this purpose.
The remainder of the paper is structured as follows.
Section \ref{sec:spare} depicts the SPARE design, highlighting its main features.
In Section \ref{sec:building_apps} we introduce the concept of \textit{generic algorithm}, showing two specific test cases; in particular in Section \ref{sec:sequences} we discuss an experiment regarding clustering of sequences, while in Section \ref{sec:graphs} we show the application of some graph matching algorithms on synthetically generated classification problems for labeled graphs.
Finally, Section \ref{sec:conclusions} concludes the paper pointing at the SPARE future directions.

\section{The SPARE Library}
\label{sec:spare}

\subsection{A Glimpse of Template Metaprogramming}
\label{sec:tmp}

A \textit{metaprogram} can be defined as a program that can manipulate itself as data or that can execute part of the job at \textit{compile time}. The language in which the metaprogram is written is called the \textit{metalanguage}, and the capability of a language to be its own metalanguage is called \textit{reflection} \citep{abrahams05C++templatemetaprogrammingconceptstoolstechniquesboostbeyond,attardi+cisternino2001}. Metaprograms work with an input language called \textit{domain language} and output language called \textit{host language}. This abstraction is provided for two main reasons: to facilitate the programmer in writing high level programs and to provide an abstraction of mechanism of program manipulation.
An example of this kind of programs is the YACC parser \citep{johnson75YACCyetanothercompilercompiler}, which can be referred to as a \textit{translator} -- a metaprogram whose domain language is different from the host language. \textit{Template metaprogramming} is a popular \textit{native language} metaprogramming technique employed by the C++ programming language, which allows to manipulate a program at compile time using \textit{templates}. The compiler translates the template generating a temporary source code that will be successively merged with the rest of the program \citep{abrahams05C++templatemetaprogrammingconceptstoolstechniquesboostbeyond,vandevoorde03C++templatescompleteguide,veldhuizen96UsingC++templatemetaprograms,stroustrup00C++ProgrammingLanguage}. Templates are an instance of the well-known \textit{generic programming} paradigm, in which algorithms are written in terms of the so-called \textit{to-be-specified-later} types \citep{stroustrup07EvolvinglanguagerealworldC++}. The exact type is 
specified as a class parameter when the algorithm is used. This mechanism permits to write reusable algorithms that abstract the data type with which they work. Other languages, such as Haskel, D, ML, Caml, XL, and modern Java, adopt a similar technique.

For example, a well-known family of classes in the C++ STL are containers \citep{becker01STLContainers,benzinger98STLcontainersbasedhashtables}. A container is a \textit{concept} that is able to store a generic type of data, providing many functionalities for manipulations, like insert, delete, and retrieval of elements. Containers are ubiquitous in every modern programming languages, such as the \textit{Java Collections} framework  \citep{long10TowardsdesignsetbasedJavacollectionsframework}. The internal low level data structures used by the specific container class are hidden to the programmer, and only high level functionalities are exposed.

A C++ concept can be defined as ``a specification of the requirements on a set of template arguments'' \citep{dosreis06SpecifyingC++concepts}. Concepts are (informal) type systems for templates, aiming to offer a unified way to define function signatures. The main issues in defining a formal type system for templates using concepts are performance requirements. A concept must be rigorous enough to be able to define the requirements of a specific group of classes and should avoid exposing too many internal specifications to the programmer. For example the concept of \textit{iterator} defines a family of classes that are able to fetch items stored in an abstract collection. The specific iteration strategy is hidden from the user point of view, and, of course, the specific data type to iterate must be defined through the template mechanism.

Many prominent C++ libraries follow this kind of specification. An example is offered by the Boost library \citep{karlsson06BeyondC++standardlibraryintroductiontoBoost,abrahams05C++templatemetaprogrammingconceptstoolstechniquesboostbeyond}, a very large and well organized collection of classes that spans between many areas, such as graphs, networking, and parallel computing. SPARE is designed in a similar way to Boost. In fact, the main design effort has been devoted to the definition of a suited collection of concepts, each one conceived in terms of classes--objects providing similar functionalities.

\subsection{Main SPARE Features}

A list of concepts and corresponding models (i.e., algorithms) available in the latest SPARE release is shown in Tables \ref{tab:spare_concepts} and \ref{tab:spare_concepts_2}.
The \textit{Dissimilarity} concept is a generalization of a distance measure, which in SPARE provides models able to process different data types. The dissimilarity measure is one of the funding blocks of many pattern recognition systems, especially when dealing with non-conventional input domains.
The \textit{Similarity} concept is practically the dual concept of the dissimilarity, and it is applicable to a broad range of data types as well. Most of the currently implemented similarity measures are well-known kernel functions widely used in pattern recognition applications.
The \textit{Evaluator} provides the abstraction of a (real-valued) function of the type $f: \mathbb{R}^d \rightarrow\mathbb{R}$; the \textit{MultiEvaluator} is the direct extension of the Evaluator concept which just generalizes the function to multidimensional outputs, i.e., $f: \mathbb{R}^d \rightarrow\mathbb{R}^m$.
By combining Evaluator and MultiEvaluator, it is possible to design well-known \textit{fuzzy inference systems} (please refer to the SPARE technical documentation for a nice ``how to'' guide).
The \textit{Representative} concept provides algorithms to calculate and update the representative of a set of patterns. In general, models implementing such a concept are applicable to different data types, while others are applicable by definition only to $\mathbb{R}^d$ data.
The \textit{Optimizer} concept includes optimization algorithms; the \textit{Environment} concept is intimately related to Optimizer since it provides specific solutions to model the so-called \textit{individuals} (i.e., the candidate solutions) in the implemented genetic algorithm.
The \textit{Representation} concept collects algorithms that aim to produce a new matrix-based representation of the input data (usually it is applicable to any input data type by means of a suited user-defined dissimilarity measure). Such a new representation could be used, for instance, to address the original problem in a suited \textit{embedding space}.
\textit{Clustering} is an auto-explicative concept collecting of course clustering algorithms, which are applicable, according to their current implementations, to virtually any input data type for which it is possible to define a dissimilarity measure.
The \textit{Supervised} concept accounts for classification systems that are trained in a supervised fashion, i.e., they are trained by making use of a labeled training set. Finally, the \textit{Unsupervised} concept provides instead classifiers which are trained in an unsupervised fashion (e.g., by means of a clustering-based model synthesis).

Table \ref{tab:spare_graph_concepts_models} provides the functionalities currently available to manage data represented as labeled graphs.
Those algorithm are presented in a separated table since they operate specifically for labeled graph data types; however, they are fully interoperable with all the other SPARE classes. Part of the herein presented concepts (see for instance Dissimilarity and Similarity) are just the specialization of the already introduced concepts for the labeled graphs data type. \textit{MatrixRepresentation} provides models to represent an input graph as a matrix (such as the adjacency matrix or the so-called transition matrix). The \textit{Operator} provides algorithms that implement some well-known (binary) operators among graphs. Finally, the \textit{Seriation} concept collects algorithmic solutions to transform an input graph into a sequence of its vertex labels.

More detailed and up-to-date technical information about the library can be retrieved from the main web site: \href{https://sourceforge.net/p/libspare/home/Spare/}{https://sourceforge.net/p/libspare/home/Spare/}.

\begin{table}[tp!]\scriptsize
\caption{List of the main SPARE concepts and implemented models -- 1/2.}
\begin{center}
\begin{tabular}{|c|c|p{7cm}|}
\hline
\rowcolor{Orange} \textbf{Concept} & \textbf{Implementation} & \textbf{Short Description}  \\

\hline
\multirow{11}{*}{Dissimilarity}
& Hamming & The standard Hamming distance for $\{0, 1\}$ data vectors. \\
\cline{2-3}
& Euclidean & The standard Euclidean distance for real valued vectors. \\
\cline{2-3}
& Minkowski & The order $p$ Minkowski distance for real valued vectors. \\
\cline{2-3}
& Dtw & The Dynamic Time Warping dissimilarity between generic data sequences. \\
\cline{2-3}
& Levenshtein & The Levenshtein edit distance. \\
\cline{2-3}
& CBMF & The Container-based Best Match First (CBMF) dissimilarity \\
\cline{2-3}
& Constant & The constant user-defined dissimilarity. \\
\cline{2-3}
& Delta & The Delta dissimilarity measure ($d(x, y)=1\Leftrightarrow x\neq y$) \\
\cline{2-3}
& DissimilarityMatrix & The dissimilarity matrix is an indexed matrix (Boost matrix) of pre-calculated dissimilarity values. It is possible to make access to the dissimilarity values according to the user-defined indices \\
\cline{2-3}
& ModuleDistance & The absolute distance (also called module) of two input objects that implement the $-$ operator \\

\hline

\multirow{8}{*}{Similarity}
& Cosine & The cosine similarity function between real-valued vectors. The function effectively computes the cosine of the angle among the two input vectors. \\
\cline{2-3}
& ExponentialKernel & The exponential kernel function. It is defined in terms of a (symmetric) dissimilarity measure, which is assumed to be able to operate on the user-defined input space. \\
\cline{2-3}
& GTSKernel & The Generalized T-Student kernel function. It is defined in terms of a (symmetric) dissimilarity measure, which is assumed to be able to operate on the user-defined input space. \\
\cline{2-3}
& HyperbolicTangentKernel & The Hyperbolic tangent kernel function between real-valued vectors. \\
\cline{2-3}
& LaplacianKernel & The Laplacian kernel function. It is defined in terms of a (symmetric) dissimilarity measure, which is assumed to be able to operate on the user-defined input space. \\
\cline{2-3}
& PolynomialKernel & The polynomial kernel function that operates between real-valued vectors. \\
\cline{2-3}
& RationalQuadraticKernel & The rational quadratic kernel function. It is defined in terms of a (symmetric) dissimilarity measure, which is assumed to be able to operate on the user-defined input space. \\
\cline{2-3}
& RBFKernel & The RBF Gaussian kernel function. It is defined in terms of a (symmetric) dissimilarity measure, which is assumed to be able to operate on the user-defined input space. \\

\hline

\multirow{3}{*}{Evaluator}
& PiecewiseLinear & A piecewise linear membership function. \\
\cline{2-3}
& Gaussian & A Monodimensional Gaussian function. \\
\cline{2-3}
& MultiGaussian & A Multivariate Gaussian function. \\
\hline

\multirow{6}{*}{MultiEvaluator} 
& FuzzyAntecedent & A fuzzy rule antecedent based on scalar fuzzy membership functions. \\
\cline{2-3}
& MultiGaussianAntecedent & A fuzzy rule antecedent based on multivariate Gaussian membership functions. \\
\cline{2-3}
& FuzzyConsequent & A fuzzy rule consequent based on scalar fuzzy membership functions. \\
\cline{2-3}
& HyperplaneConsequent & A fuzzy rule consequent computed as a linear polinomial function of rule's inputs. \\
\cline{2-3}
& FuzzyRuleBlock & A block of fuzzy logic rules. \\
\hline

\multirow{4}{*}{Representative}
& Centroid & Mean vector based representative of a cluster of real valued vectors. The point to cluster dissimilarity is user defined. \\
\cline{2-3}
& Mahalanobis & Representative based on mean vector and covariance matrix for a cluster of real valued vectors. The point to cluster dissimilarity is computed as the Mahalanobis distance. \\
\cline{2-3}
& MinSod & Representative of a generic cluster of objects, based on the cluster sample which minimizes the Sum Of Distances from the other samples. The point to cluster dissimilarity is user defined. \\
\cline{2-3}
& FuzzyMinSod & Same as MinSod, but it offers the functionality of computing the membership function of the considered objects. The membership function is calculated according to an Evaluator function that elaborates the distance with respect to the current MinSOD element. \\
\cline{2-3}
& PFuzzyMinSod & Same as FuzzyMinSod, but is characterized by a different replacement policy. The elements are replaced according to a probability distribution computed considering the dissimilarity values with respect to the MinSOD element. Elements with lower dissimilarity values have lower probability of being replaced. \\
\cline{2-3}
& Rlse & Representative of a cluster of input-output pairs, where inputs are real valued vectors and output are scalar real values, based on a hyperplane approximating the points in the cluster. The point to cluster dissimilarity is computed as the hyperplane approximation error. \\
\cline{2-3}
& FuzzyHyperbox & Model of a fuzzy set-based hyperbox. \\

\hline
\end{tabular}
\end{center}
\label{tab:spare_concepts}
\end{table}

\newpage

\begin{table}[tp!]\scriptsize
\caption{(Continuation of Table \ref{tab:spare_concepts}) List of the main SPARE concepts and implemented models -- 2/2.}
\begin{center}
\begin{tabular}{|c|c|p{7cm}|}
\hline
\rowcolor{Orange} \textbf{Concept} & \textbf{Implementation} & \textbf{Short Description}  \\

\hline
\multirow{2}{*}{Optimizer} & GeneticAlgorithm & An implementation of a classic genetic algorithm. \\
\cline{2-3}
& PGeneticAlgorithm & Same genetic algorithm implementation but with multi-threading capability. \\

\hline
Environment & DiscreteCode & An environment agent to be used with the genetic algorithm implementation, where the candidate solution (individual) is coded as an integer-valued vector. Classical crossover and mutation operators are provided within the class. \\

\hline
Representation & DissimilarityRepr & The dissimilarity representation of an input set with respect to a representation set. It is generalized according to a specific type of dissimilarity measure. \\

\hline

\multirow{3}{*}{Clustering} & Kmeans & The standard \textit{k}-means algorithm generalized for what concerns the cluster representatives initialization scheme (we provide different initialization algorithms that can be also user-defined). \\
\cline{2-3}
& Bsas & The Basic Sequential Algorithmic Scheme clustering algorithm. Fast linear-time cluster generation rule. \\
\cline{2-3}
& MTBsas & Same as Bsas but with multi-threading capability. \\

\hline

\multirow{8}{*}{Supervised} & KnnApprox & A version of the \textit{k} nearest neighborhood algorithm for function approximation problems. \\
\cline{2-3}
& KnnClass & The well-known \textit{k} nearest neighborhood algorithm for classification problems. \\
\cline{2-3}
& MTKnnClass & Same as KnnClass but with multi-threading capability. \\
\cline{2-3}
& SvmApprox & A version of the SVM algorithm for function approximation problems (uses \href{http://www.csie.ntu.edu.tw/~cjlin/libsvm/}{libsvm} as core engine). \\
\cline{2-3}
& SvmClass & The well-known SVM algorithm for classification problems (uses \href{http://www.csie.ntu.edu.tw/~cjlin/libsvm/}{libsvm} as core engine). \\
\cline{2-3}
& Scbc & A supervised clustering-based classification system for general user-defined data types. \\
\cline{2-3}
& MMNet & A neurofuzzy Min-Max network trained with the ARC/PARC algorithms \cite{rizzi2002}. \\
\cline{2-3}
& DlibSvmClass & A wrapper of the \href{http://dlib.net/ml.html}{Dlib} implementation of the kernelized C-SVM and $\nu$-SVM. It accepts a user-defined kernel function which can be generalized to virtually any data type. \\
\cline{2-3}
& DlibSvmApprox & A wrapper of the \href{http://dlib.net/ml.html}{Dlib} implementation of the kernelized $\epsilon$-SVR. It accepts a user-defined kernel function, which can be generalized to virtually any data type. \\

\hline

\multirow{2}{*}{Unsupervised} & Rlrpa & A reinforcement learning based classification algorithm for user-defined data types. \\
\cline{2-3}
& Ucbc & The same as Scbc but synthesizes the classification model in an unsupervised fashion. \\

\hline
\end{tabular}
\end{center}
\label{tab:spare_concepts_2}
\end{table}

\begin{table}[tp]\scriptsize
\caption{List of SPARE concepts and currently implemented models for graph-based pattern analysis.}
\begin{center}
\begin{tabular}{|c|c|p{7cm}|}
\hline
\rowcolor{Orange} \textbf{Concept} & \textbf{Implementation} & \textbf{Short Description}  \\
\hline
\multirow{5}{*}{Dissimilarity}
& PGED & The Perfect GED class implements the GED algorithm that solves the GED problem optimally. \\
\cline{2-3}
& BMF & The weighted (vertex) Best Match First graph dissimilarity measure implements a vertex assignment solution following a greedy heuristic. \\
\cline{2-3}
& sBMF & The Shuffled BMF graph dissimilarity implements the same BMF matching scheme. The algorithm allows the user to select a number $s$, which is the number of shuffles to apply to the vertex set of the input graphs. Consequently, the sBMF algorithm consists in executing the BMF algorithm $s$ times, returning the lower overall dissimilarity value. \\
\cline{2-3}
& TWEC & The Triple Weighted Error Correcting (TWEC) scheme is basically the normalized Best Match First algorithm, yielding values within the $[0, 1]$ range. \\
\cline{2-3}
& HGED & This GED is based on the Hungarian algorithm for optimally solving the assignment problem of the vertices. The procedure returns dissimilarity values within the $[0, 1]$ range. \\
\hline

Similarity & GraphCoverage & The Graph Coverage similarity algorithm between arbitrarily labeled graphs. The algorithm was presented here \cite{livi2012gc,livi2012_pgm}, and it is based on the tensor product computation. \\
\hline

\multirow{9}{*}{MatrixRepresentation}
& Adjacency & The standard adjacency matrix representation of a graph. \\
\cline{2-3}
& Eigenvectors & The eigenvectors-based matrix representation for graphs. This class uses the Eigen library. \\
\cline{2-3}
& InducedTransition & The induced transition matrix is a standard transition matrix computed using the Euclidean norm of the numeric vectors of the edges of the graph. \\
\cline{2-3}
& Laplacian & The Laplacian matrix representation of a graph. \\
\cline{2-3}
& NormalizedLaplacian & The normalized Laplacian matrix representation of a graph. \\
\cline{2-3}
& SymTransition & The symmetric transition matrix representation of a graph. \\
\cline{2-3}
& Transition & The standard transition matrix representation of a graph. \\
\cline{2-3}
& WTransition & The transition matrix of a graph using the real-valued labels (the weights) of the graph's edges to induce the transition probabilities. \\
\cline{2-3}
& HeatMatrix & The heat matrix is elaborated from the normalized graph Laplacian via the exponentiation of the eigenvalues at a given time $t$. \\
\hline

\multirow{3}{*}{Operator}
& TensorProduct & The standard tensor product (also called Kroneker product or direct product) operator between two labeled graphs. The algorithm yields an output graph described in terms of weighted adjacency matrix. \\
\cline{2-3}
& SymTensorProduct & The tensor product computed assuming symmetric matrices representations (that is, undirected graphs). \\
\cline{2-3}
& MCTensorProduct & The multicore implementation of the standard tensor product operator. \\
\hline

\multirow{1}{*}{Seriation}
& EigenSeriation & The Eigenvectors-based seriation algorithm implemented according to \cite{robleskelly+hancock2005}. The algorithm produces an output sequence containing an ordering of the graph vertex labels according to the first eigenvector information of the graph matrix representation. \\
\hline

\end{tabular}
\end{center}
\label{tab:spare_graph_concepts_models}
\end{table}

\clearpage
\section{Building Pattern Recognition Applications with SPARE}
\label{sec:building_apps}

In this section, we first introduce the idea of generalizing pattern recognition mechanisms. Successively, we consolidate this aspect by providing two different specific application instances. In particular, Section \ref{sec:clustering_go} discusses, in terms of clustering, two important concepts in this scenario: the dissimilarity measure and a set (cluster) representation technique in the non-geometric case. In Section \ref{sec:sequences} we show the first application involving clustering of real-valued multidimensional sequences, while Section \ref{sec:graphs} shows a benchmark of some graph matching algorithms in the context of classification.
In both sections, we report some (minimal) technical details concerning how to combine some of the key SPARE classes needed in those tests. We refer the interested reader to the technical documentation and the practical examples provided within the SPARE installation package for further and more specific details.

\subsection{Clustering of a Set of Generic Objects}
\label{sec:clustering_go}

Cluster analysis, or clustering, is one of the classic problems in pattern recognition. The problem focuses on the task of grouping a set of objects in such a way that objects in the same group (called cluster) are more \textit{similar}, or \textit{compatible}, to each other with respect to those in other clusters.
In spite of the availability of a large number of clustering algorithms, and their proved success in a number of different application domains, clustering remains a difficult and ill-posed problem \cite{Jain:2010:DCY:1755267.1755654}. This fact can be attributed to the inherent vagueness in the definition of a cluster, which magnifies the difficulty of defining an appropriate similarity measure and an effective optimization procedure to guide the clustering process.

The problem of grouping a set of elements into a predefined number $k$ of different clusters is called \emph{k clustering problem}.
Given a finite input set $\mathcal{X}=\{x_{1},...,x_{n}\}$, where $x_{i}\in\mathbb{R}^{n}, i=1,...,n$, a $k$ clustering problem consists in finding a \emph{partition} $\mathcal{P}=\{C_{1},...,C_{k}\}$ of this set in a number $k\leq n$ of clusters, with $\mathcal{X}=\bigcup_{i=1}^{k} \mathcal{C}_{i}$ and $\mathcal{C}_{i}\cap\mathcal{C}_{j}=\emptyset,\forall i,j, i\neq j$, so that for a given metric in $\mathbb{R}^n$, ``close'' patterns will fall in the same cluster, while ``sufficiently far'' patterns will belong to different clusters. In order to solve this problem, the celebrated \textit{k}-means algorithm \cite{theodoridis2006pattern} tries to minimize the within-cluster sum of squares
\begin{equation}
\displaystyle\argmin_{\mathcal{P}} \displaystyle\sum_{i=1}^{k}\displaystyle\sum_{x_{j}\in\mathcal{C}_{i}} \lVert x_{j}-\mu_{i} \rVert^{2} ,
\end{equation}
where $\mu_{i}$ is the \textit{mean} vector of the \textit{i}-th cluster $\mathcal{C}_{i}$.
The number $k$ is called the \textit{order} of the partition and it is defined a priori.
\textit{k}-means operates as a two-stage iterative \textit{heuristic} algorithm that is based on the assignment of each input pattern $x_{i}$ to the cluster with the closest mean $\mu_{j}$. After the \textit{assignment} phase follows the \textit{update} phase, consisting in updating all clusters' means $\mu_j$. The two phases are iterated until a predefined stop criterion is true.

In order to define clustering algorithms not necessarily dealing with a metric space (e.g., an Euclidean space), we observe some general (\emph{meta}) elements which play a fundamental role in the clustering process. Each cluster is modeled by a \textit{representative}, i.e., an element not necessarily belonging to the input data space, which is able to characterize the entire set of patterns in the cluster. Another important concept is the \textit{dissimilarity measure} used to compute the distance between patterns and representative elements. It is easy to understand that these two definitions of representative of a cluster and of dissimilarity measure strictly depend on the data type characterizing the problem at hand. For example, if $\mathcal{X}=\mathcal{G}$, where $\mathcal{G}$ is a set of graphs, the problem of deriving a representative graph is known as the \textit{median graph} identification \cite{jiang+mnger+bunke2001,ferrer10Generalizedmediangraphcomputationbymeansgraphembeddingvectorspaces,
bardaji+ferrer+sanfeliu2010}.
By abstracting the input domain, it is possible to define a \textit{generic} k-means algorithm that uses a generic representative mechanism, which is in turn configured with a generic dissimilarity measure. That is, it is possible to abstract the algorithm itself obtaining thus a \textit{meta-algorithm} that will work on many different problems, but preserving the main behavior (this is the very core of template metaprogramming). The same generalization could be applied to many pattern recognition problems, such as classification and function approximation.

Algorithm \ref{alg:kmeans} shows the pseudo-code of the generic \textit{k}-means algorithm over an input set $\mathcal{X}\subset\mathcal{U}$ ($\mathcal{U}$ is a given input domain), which is configured to operate by means of $k$ set representatives $\{\mu_{1},...,\mu_{k}\}$. The distance among two input elements is calculated through the dissimilarity measure $d: \mathcal{U}\times\mathcal{U}\rightarrow\mathbb{R}^{+}$.
The pseudo-code of Algorithm \ref{alg:kmeans} shows that the dissimilarity evaluation between a sample $x_{j}$ and a cluster $\mathcal{C}_{i}$ can be computed by comparing the sample with the representative of the cluster, namely $\mu_{i}$, using the dissimilarity function $d(\cdot,\cdot)$ (line 6), defined over the generic space $\mathcal{U}$. The update mechanism of each representative is carried out by a generic operator $R(\cdot)$ (line 8), able to derive the representative of the modified cluster $\mathcal{C}_{i}$. At each step $t$, a new partition $\mathcal{P}^{t}$ is induced (line 9), until the stop condition is reached (line 10), which is usually defined as a combination of a maximum number of iterations and other criteria.

\begin{algorithm}\footnotesize
\caption{Generic \textit{k}-means.}
\label{alg:kmeans}
\begin{algorithmic}[1]
\REQUIRE A generic finite input set $\mathcal{X}=\{x_{1},...,x_{n}\}$, the order $k$ of the clustering, a dissimilarity function $d: \mathcal{U}\times\mathcal{U}\rightarrow\mathbb{R}^{+}_{0}$, the cluster representatives $\{\mu_{1},...,\mu_{k}\}$
\ENSURE A partition $\mathcal{P}=\{\mathcal{C}_{1},...,\mathcal{C}_{k}\}$
\STATE Initialize every representative $\mu_{i}, i=1,...,k$
\STATE $t=0, \mathcal{P}=\emptyset$
\LOOP
\STATE $t+=1$
\STATE Assignment Step: assign each sample $x_{j}$ to the cluster with the closest representative
\STATE $C_i^{(t)} = \left\{ x_{j} : d(x_{j}, \mu_{i}) \leq d(x_{j}, \mu_{h}) \text{ for all }h=1,\ldots,k \right\}$
\STATE Update Step: update the representatives
\STATE $\mu^{(t+1)}_i=R(\mathcal{C}_{i}),\ i=1\rightarrow k$
\STATE Update the partition with the modified clusters: \\ $\mathcal{P}^{t}=\{\mathcal{C}_{1},...,\mathcal{C}_{k}\}$
\IF{STOP}
\RETURN $\mathcal{P}^{t}$
\ENDIF
\ENDLOOP
\end{algorithmic}
\end{algorithm}

\subsubsection{The Dissimilarity Approach in Pattern Recognition}
\label{sec:dissimilarity}

Pattern recognition mechanisms are deeply grounded on the possibility of \textit{distinguishing} among the input patterns, in either a generative of discriminative setting \cite{gm_survey,pkekalska+duin2005}. By departing from the common Euclidean space, we therefore face problems related to the difficulty of defining an effective dissimilarity measure $d:\mathcal{U}\times\mathcal{U}\rightarrow\mathbb{R}^+$ on the input space $\mathcal{U}$. A dissimilarity is basically the generalization of a metric distance, which may satisfy fewer requirements. When the input space has no trivial geometry, such as in the case of labeled graphs \cite{gm_survey} and sequences \cite{grapsec_ijcnn_2013,t2apdiss__ifsanafips2013,t2vsdiss__ifsanafips2013}, the definition of such a dissimilarity measure can become a crucial task. Reasonable dissimilarity measures are usually positive and definite functions; however most of the times such algorithms are also symmetric. The triangle inequality instead is not easily 
achievable in the general case. As pointed out by many authors (see e.g. \cite{gm_survey,pkekalska+duin2005,nonmetric_informative__2006,odse}) from the pattern recognition viewpoint, adhering to the metric requirements does not necessarily imply better performances in terms of pattern recognition problems.

Closely related to the concept of dissimilarity is the one of similarity, which however must be interpreted as the dual concept (it has the dual meaning). The two concepts are intimately related. In fact, it is always possible to derive a similarity measure from the corresponding dissimilarity, and vice versa. A particularly interesting type of similarity measures is provided by \textit{positive definite} kernel functions. A positive definite (pd) kernel function is a symmetric function of the type $k:\mathcal{U}\times\mathcal{U}\rightarrow\mathbb{R}^+$. Well-known pd kernels in pattern recognition are the RBF Gaussian, the Laplacian, the polynomial, and linear kernel \cite{schoelkopf+smola2002}.

Dissimilarity and pd kernel functions play a pivotal role in implementing generic pattern recognition applications that can operate on user-defined input spaces. SPARE provides the implementation of different dissimilarity measures operating on vectors of real values, sequences of generic objects, labeled graphs, and set of generic objects. Moreover, it includes the implementation of many pd kernel functions. A particular type of kernel functions is the one that is defined in terms of a dissimilarity (such as the RBF Gaussian). In this case, pd kernels are easily extended to different input domains by means of the interoperation between the dissimilarity and similarity concepts.

\subsubsection{The MinSOD Cluster Representative}
\label{sec:minsod}

Modeling a finite input set $\mathcal{X}=\{x_{1},...,x_{n}\}$ of generic objects is an important aspect of pattern recognition applications. For example, if $\mathcal{X}\subset\mathbb{R}^{n}$ the representative of such a set may be simply the \textit{centroid}, that is, a vector $\underline{\mathbf{x}}^{*}=\frac{1}{n}\sum_{i} \underline{\mathbf{x}}_{i}$, with $\underline{\mathbf{x}}^{*}\in\mathbb{R}^{n}$, which does not necessary belong to the input set $\mathcal{X}$.
The problem of modeling a set of objects can be generalized to virtually any input domain where it is possible to define a dissimilarity measure $d(\cdot,\cdot)$. Formally, the representative element can be computed \cite{delvescovo+livi+rizzi+frattalemascioli2011} as follows:
\begin{equation}
\label{eq:minsod}
x^{*}=\displaystyle\argmin_{x_{i}\in\mathcal{X}} \sum_{x_j\in\mathcal{X}, x_j\neq x_i} d(x_{j}, x_{i}).
\end{equation}

The \textit{set mean} in this case is taken as the element of the set $\mathcal{X}$ that minimizes the sum of the distances (MinSOD) between the element itself and the other elements in the set. To obtain the MinSOD we need to execute a quadratic number of distance calculations. That is, the overall complexity depends in turn on the computational complexity of $d(\cdot,\cdot)$, which, especially when dealing with complex data structures \cite{gm_survey}, is usually polynomial. If $\mathcal{X}$ is a large set, the evaluation of the distance between every objects of $\mathcal{X}$ become prohibitive and some approximation mechanism must be taken into account.

In SPARE the MinSOD determination is achieved through a specific class, named \emph{MinSod}, which models the \textit{Representative} concept.
The MinSod object performs an important computational speed up by only determining the MinSOD representative inside a reduced pool of samples, instead of the whole set of the input samples inserted in the cluster so far. The reduced pool of samples is called the \textit{cache} and its size is a relevant parameter for the user to setup. In order to work with a reduced set of samples, a \textit{replacement policy} has to be defined to discard some samples from the pool as new samples are inserted in the set. In SPARE, the MinSOD has been implemented according to different replacement policies, which are specialized for different purposes. The standard MinSOD replacement policy is simple but efficient in practice. Each new element is inserted since the defined cache size is reached. Then, for each new sample an old one must be discarded. The old sample to discard is chosen as follows: two samples are chosen with uniform probability, and the one farthest away from the actual MinSOD representative is discarded.
We refer the interested reader to the SPARE technical documentation for more details.
In Algorithm \ref{alg:minsod} is shown the pseudo-code of the behavior of the MinSod class of SPARE.

 \begin{algorithm}\footnotesize
\caption{Standard MinSOD representative and related cache replacement.}
\label{alg:minsod}
\begin{algorithmic}[1]
\REQUIRE A generic finite input set $\mathcal{X}=\{x_{1},...,x_{n}\}$, the size $M$ of the cache $\mathcal{C}$, a dissimilarity function $d: \mathcal{X}\times\mathcal{X}\rightarrow\mathbb{R}^{+}_{0}$
\ENSURE The MinSOD determination $\hat x \in \mathcal{X}$
\STATE $\mathcal{C}=\emptyset$
\FORALL{$x_i\in\mathcal{X}$}
\IF{$|\mathcal{C}| < M$}
\STATE $\mathcal{C}=\mathcal{C}\cup\{x_i\}$
\ELSE
\STATE Select two elements, namely $\chi_1$ and $\chi_2$ from $\mathcal{C}$ with uniform probability
\IF{$d(\chi_1, \hat x) \geq d(\chi_2, \hat x)$}
\STATE $\mathcal{C}=\mathcal{C} \setminus \{\chi_1\}$
\ELSE
\STATE $\mathcal{C}=\mathcal{C} \setminus \{\chi_2\}$
\ENDIF
\STATE $\mathcal{C}=\mathcal{C}\cup\{x_i\}$
\ENDIF
\STATE Determine $\hat x=\mathrm{MinSOD}(\mathcal{C})$ according to Equation \ref{eq:minsod}
\ENDFOR
\end{algorithmic}
\end{algorithm}

\subsection{Sequence Clustering and MinSOD Representative Evaluation}
\label{sec:sequences}

In this section we deal with an input set containing real-valued multidimensional sequences.
In particular, we will conceive a clustering problem involving the aforementioned standard MinSOD cluster model, testing its robustness when varying the cache size parameter. Since we deal with input patterns that are sequences of real-valued vectors, we will equip the MinSOD representative with a dissimilarity measure elaborated from the well-known Dynamic Time Warping (DTW) matching algorithm \cite{izakian2015fuzzy,grapsec_ijcnn_2013,t2apdiss__ifsanafips2013,t2vsdiss__ifsanafips2013,sakoe1978,Nakamura__ts,Berndt_Clifford__1996}.
The DTW follows a global alignment scheme, which has been proposed to align sequences of arbitrarily-defined objects belonging to a set $\mathcal{A}$ (e.g., sequences of characters, sequences of real-valued vectors, or even sequences of complex data structures). The key component of the algorithm is the core dissimilarity measure, $d_{\mathcal{A}}: \mathcal{A}\times\mathcal{A}\rightarrow\mathbb{R}^{+}$, which quantifies the cost of comparing two input objects. In the herein discussed problem, the objects are 5-dimensional real-valued vectors, and the core dissimilarity measure of the DTW is the Euclidean metric.

The considered clustering problem is defined as a two-class classification problem, with each class being modeled by a single cluster and where each object of the sequence is a 5-dimensional real-valued vector extracted from a 5-dimensional Gaussian distribution.
Each class contain 150 sequences (for a total of 300 sequences for each problem instance), and it is characterized by a specific Gaussian distribution with a certain mean and (spherical) covariance matrix. By fixing the mean value of the two 5-dimensional Gaussian distributions, we can control the class overlap by increasing the variance parameter of the respective covariance matrices.
The objective is to group sequences that pertain to the same generating class, that is, are generated from the same Gaussian distribution. This is in all respects an unsupervised learning task. Once the partition is obtained, a simple quality index which assumes values in $[0, 1]$ is evaluated for each computed cluster. The quality index takes into account the \textit{purity} of the cluster in terms of the known class labels; it is defined by the number of samples belonging to the majority class in the cluster, divided by the total cardinality of the cluster. An index value of one indicates a pure cluster, composed by samples of the same class only. An index value of 0.5 indicates a cluster which contains half of the samples from one class, half from the other. The overall quality index is taken as the worst quality index over the two clusters. 
We show the results obtained over two different tests, with the second problem representing a more difficult clustering problem (higher 
overlap of the Gaussian distributions) with respect to the first one. Our aim is to observe how, and if, reducing the cache size of the MinSOD affects the established performance measure.
To this end, the cache size is decremented, for both tests, starting from a value of 50 and going down to 1, with a unit decrement step. For each size of the cache, the same test is carried out with five different random seeds. The results for the quality index are eventually taken as the average over these runs; the worst and better case results are also considered at each cache size.

\subsubsection{Code Snippets}

The practical SPARE configuration of the DTW template class for processing such input patterns looks like:
\begin{lstlisting}[frame=tb,caption=DTW configuration for processing sequences of real-valued vectors.,label=dtw]
typedef std::vector<double> SequenceObjectType;
typedef std::vector<SequenceObjectType> SequenceType;
typedef Euclidean SequenceOjectDissimilarityType;
typedef Dtw<SequenceOjectDissimilarityType> SequenceDissimilarityType;
\end{lstlisting}

Once we have defined the dissimilarity measure for the considered type of sequences, we also need to define the MinSOD cluster model type, and finally the clustering algorithm, which operates by using such cluster models. We adopt the \textit{k}-means implementation of SPARE as the clustering algorithm. The SPARE code to set up such classes via type definitions is:
\begin{lstlisting}[frame=tb,caption=Configuration of the MinSOD and clustering algorithm.,label=minsod_clustering]
typedef MinSod<SequenceType, SequenceDissimilarityType> MinSodType;
typedef FirstK KmeansInitializerType;
typedef Kmeans<MinSodType, KmeansInitializerType> KmeansAlgoType;
\end{lstlisting}

The herein prepared \textit{k}-means type is then instantiated as a runtime object and used via the main method called \textit{Process}, by using the standard STL container-based data passing interface. In Listing \ref{lst:clustering} the call to the \textit{Process} method takes as the first two arguments the starting and ending iterators delimiting the patterns to be analyzed into the container (assumed to be stored into the \textit{DataSet} container). The third argument, i.e., the container called \textit{Labels}, accounts for the output labels describing the calculated cluster associations of the input patterns.
\begin{lstlisting}[frame=tb,caption=Processing of the input data.,label=lst:clustering]
std::vector<SequenceType> DataSet;
std::vector<NaturalType> Labels;
...
KmeansAlgoType clusteringAlgorithm;
...
clusteringAlgorithm.Process(DataSet.begin(), 
                                    DataSet.end(), Labels.begin());
\end{lstlisting}

The user can then use those output labels, or the set of calculated cluster representatives accessible via the clustering algorithm object, for its own practical purposes.

It is very easy to understand that such a setup can be very easily generalized to any user-defined data type. Moreover, switching among the possible cluster models and pattern dissimilarity measures is totally independent from the clustering algorithm functionality, which simplifies the practical development of generic and reusable pattern recognition systems.

\subsubsection{Results}

\begin{figure}[ht!]
\centering
\subfigure[]{
    \includegraphics[viewport=0 0 348 243,scale=0.55,keepaspectratio=true]{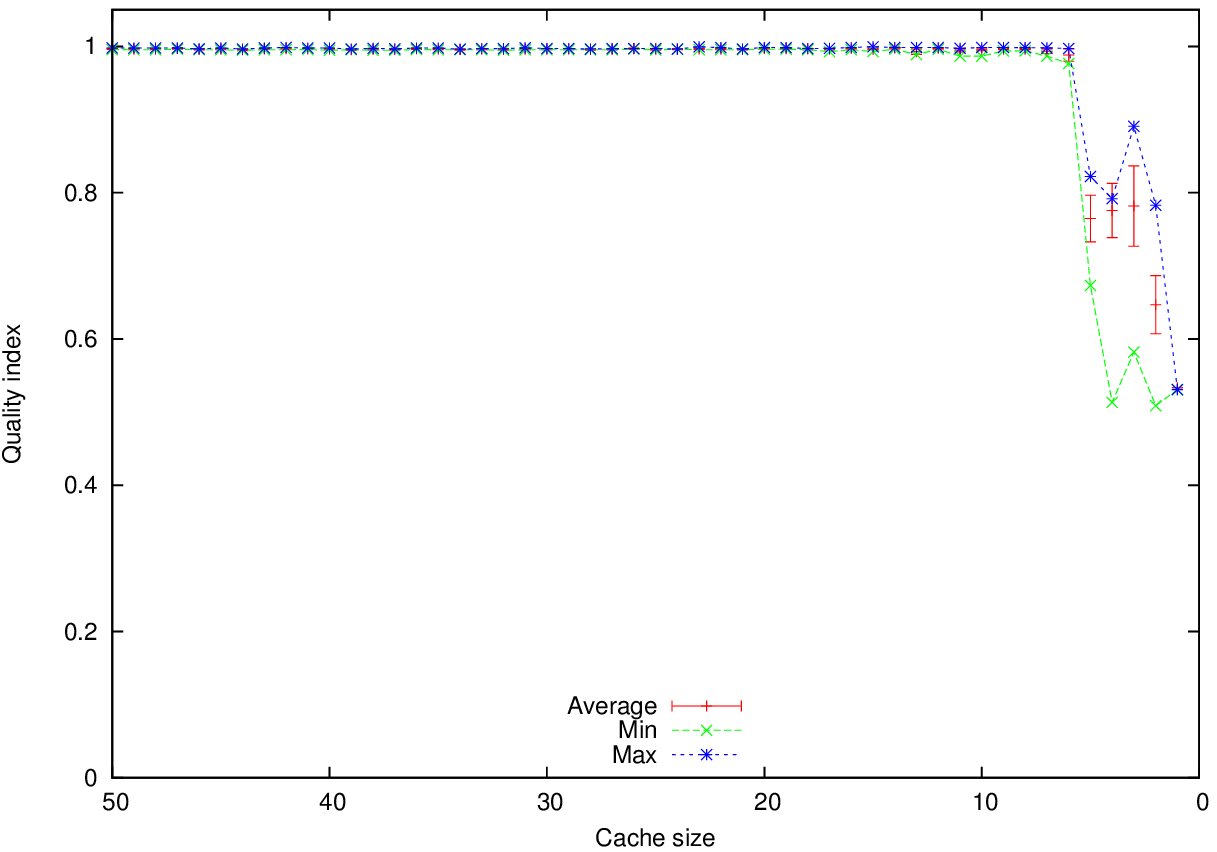}
 \label{fig:seq1}
 }
~
\subfigure[]{
 \includegraphics[viewport=0 0 348 243,scale=0.55,keepaspectratio=true]{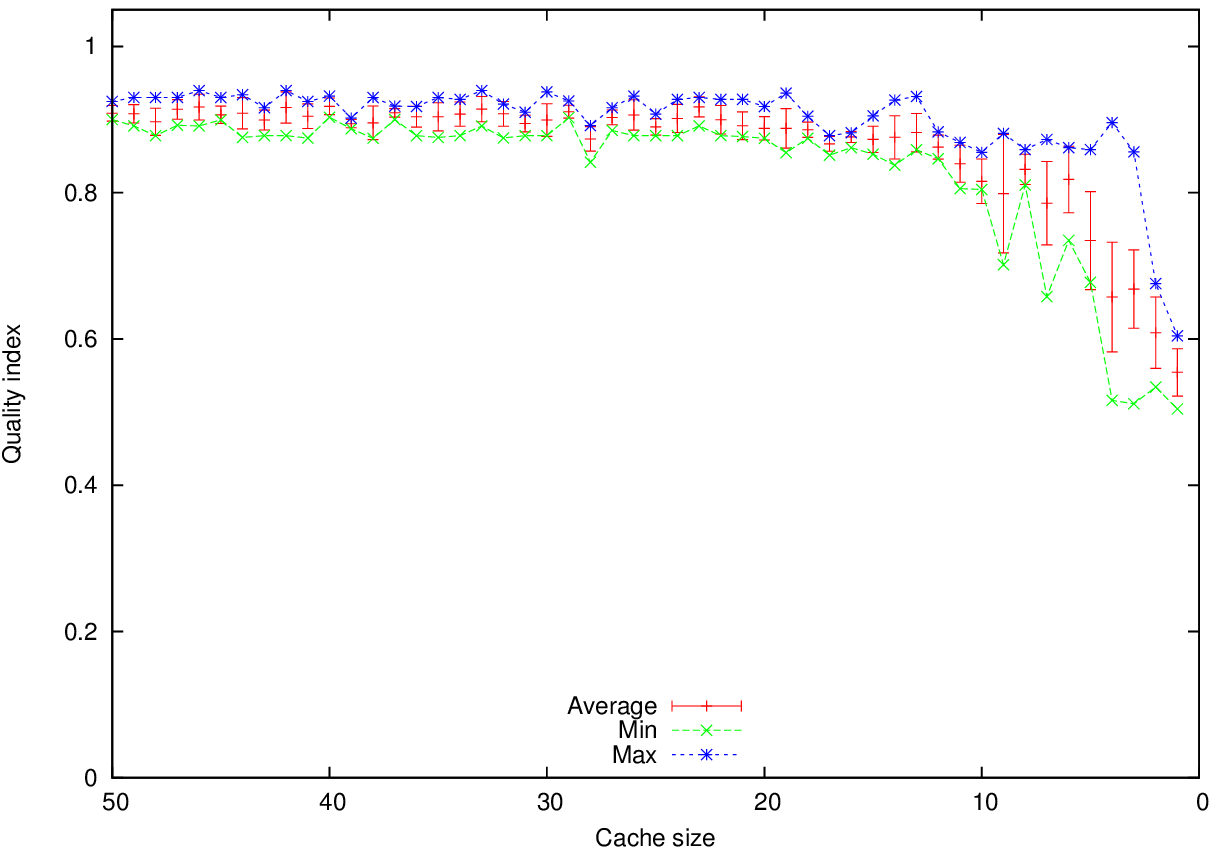}
 \label{fig:seq2}
}
\caption{The average performance results (quality index vs cache size) are reported together with the standard deviation (vertical bars). \ref{fig:seq1} shows the results for the first test, while \ref{fig:seq2} for the second one.}
\label{fig:seq_all}
\end{figure}

The results of first test are shown in Figure \ref{fig:seq1}. It is possible to observe a clear breakdown at a cache size of dimension 5. For higher values the results are very stable and nearly optimal (i.e., close to one), meaning that the MinSOD cluster models effectively the input classes.
Also in the second test the cache size of MinSOD varies from 50 to 1, with a decrement step of 1; the same considerations about the random seed initialization hold also in this case. The results for this second test are shown in Figure \ref{fig:seq2}. In this case, since the problem is more difficult, performances start to seriously deteriorate from a cache size around 10.
The performance tests show that the implemented MinSOD algorithm is able to model effectively the input set, with a correct behavior even when reducing the cache size.
In fact, before reaching a very low critical threshold for the cache size, the size is quite irrelevant for the cluster modeling purpose. A very good stability is observed with respect to the randomized replacement policy. This stability is observed along the whole interval of the valid settings of the cache size, while strong variations arise only in the region where the cache size is too low to guarantee a correct coverage of the input set.

\subsection{Benchmarking Graph Matching Algorithms in Classification Problems}
\label{sec:graphs}

As a further example of the versatility of the SPARE design, in this section we discuss the application of some graph matching algorithms implemented in SPARE, on different classification problem instances defined on synthetically generated labeled graphs. Graph matching algorithms play a key role in the design of graph-based pattern recognition systems \cite{gralg_2012,odse,gm_survey,odse2__arxiv,odse2_ijcnn_2013,gao+xiao+tao+li2010,Noma20121159,Bengoetxea20022867,dyngraph_ijcnn_2013}, since they provide the basic mechanism to measure the graph dissimilarity value.

Each synthetic dataset has been generated using the same Markov chains based method described by \citet{livi+delvescovo+rizzi_seriation+gradis}. We have conceived 15 different two-classes classification problem instances, which are characterized by a decreasing level of difficulty. Each problem is defined by a training, a validation, and a test set, containing 300 graphs each of order 30 and a variable size between 40 and 75.
The hardness of the problem has been controlled generating the labels of both vertices and edges as numeric vectors falling in $[0, 1]^5$. Each random vector has been constructed sampling numbers from two different Gaussian distributions (i.e., one for each class), distributed with suitably chosen means and standard deviations.
In addition, the stochastic generation process contributes in generating graphs with a randomized topology.

In this experiment, we have considered four different graph classification systems, which are based on a classifier relying on the \textit{k}-NN rule. The \textit{k}-NN rule offers a perfect example to employ the concept of dissimilarity operating directly on the input space. In fact, the neighborhood selection, as well as the classification rule, is carried out completely on the base of the outcomes yielded by the adopted dissimilarity measure. Therefore, since we deal with labeled graphs, we test three \textit{k}-NN based graph classification systems that operate directly on the input space $\mathcal{G}$ of labeled graphs by means of suited inexact graph matching algorithms \cite{gm_survey}.
Two of the considered algorithms adopt the BMF-GED algorithm using the 6 weights (PD6W) and the 3 weights (TWEC) schemes for the overall edit distance computation \cite{gm_survey}. The third one instead is a graph kernel function called \textit{Graph Coverage} (GC) \cite{livi2012gc,livi2012_pgm}. Finally, a graph seriation-based system equipped with the DTW distance (Seriation), tailored for sequences of five-dimensional real-valued vectors, is considered \cite{livi+delvescovo+rizzi_seriation+gradis}.

\subsubsection{Code Snippets}

In Listing \ref{knn_setup}, we briefly summarize the main SPARE classes involved in the configuration of such a graph classifier in the TWEC case; other cases follow straightforwardly.
The first thing to define is the type of considered labeled graph. Since we deal with graphs having vertex and edge labels defined as 5-dimensional real-valued vectors, the SPARE graph type to be considered is the ``VectorGraph'' type. Such a graph structure is provided with the SPARE graph package; however, the user can define virtually any type of labeled graph. The other involved types define the graph dissimilarity (e.g., the TWEC graph matching in this example) and the \textit{k}-NN classifier. Such a classifier is able to operate directly on the considered labeled graph domain by means of the direct evaluations of the graph dissimilarity measure.
\begin{lstlisting}[frame=tb,caption=\textit{k}-NN Classifier equipped with the TWEC graph dissimilarity measure.,label=knn_setup]
typedef VectorGraph LabeledGraphType;
typedef string LabelType;
typedef Euclidean VertexLabelsDissimilarityType;
typedef Euclidean EdgeLabelsDissimilarityType;
typedef TWEC<VertexLabelsDissimilarityType, 
                EdgeLabelsDissimilarityType> GraphDissimilarityType;
typedef KnnClass<LabeledGraphType, GraphDissimilarityType, 
                                          LabelType> ClassifierType;
\end{lstlisting}

By making use of the data types defined in the previous listing, we can now allocate and configure the graph classifier.
After the object allocation, we can train the classifier by using the \textit{Learn} method (note however that the herein considered \textit{k}-NN does not have a proper model synthesis stage; however this scheme applies to any SPARE classifiers which do also synthesize a classification model) and then test the synthesized model by calling the \textit{Process} method, as shown in Listing \ref{knn}.
The \textit{Learn} method accepts three arguments: the first two are the iterators delimiting the input training graphs into the container, while the third one is an iterator pointing at the first class label corresponding to the first training graph.
The \textit{Process} method works analogously, although the method writes in the container referenced by the output label iterator (third argument) the calculated class labels for the test graphs. The user is then in charge of defining a meaningful performance (i.e., error or accuracy) measure with respect to the effective test graph class labels.
\begin{lstlisting}[frame=tb,caption=Learning and testing the graph classifier.,label=knn]
std::vector<LabeledGraphType> trainGraphs, testGraphs;
std::vector<LabelType> trainLabels, testLabels, outLabels;
...
ClassifierType graphClassifier;
...
graphClassifier.Learn(trainGraphs.begin(), trainGraphs.end(), 
                                               trainLabels.begin());
//output labels allocation
outLabels.assign(testGraphs.size(), "undetermined");
graphClassifier.Process(testGraphs.begin(), testGraphs.end(), 
                                                 outLabels.begin());
\end{lstlisting}

Again, it is very easy to understand that the generalization of the \textit{k}-NN rule on other data types is very straightforward, requiring few practical changes in herein presented setup.

\subsubsection{Results}

Tests have been performed using three values for the \textit{k} parameter defining the considered nearest neighbors in the \textit{k}-NN classification rule: namely $1, 3$, and $5$.
Since the considered graph matching algorithms are sensible to the setting of the weighting parameters, we optimize those parameters by a cross validation procedure. The global optimization is performed by means of the genetic algorithm implemented in SPARE. The fitness function guiding the optimization is thus defined considering directly the recognition rate achieved on the validation set. The setup and integration of the genetic algorithm is omitted here for the sake of brevity; the interested reader can however refer to the ``GKnn.cpp'' example provided with the SPARE package.

\begin{figure}[ht!]
\centering

\subfigure[Results with $k=1$.]{
    \includegraphics[viewport=0 0 346 243,scale=0.41,keepaspectratio=true]{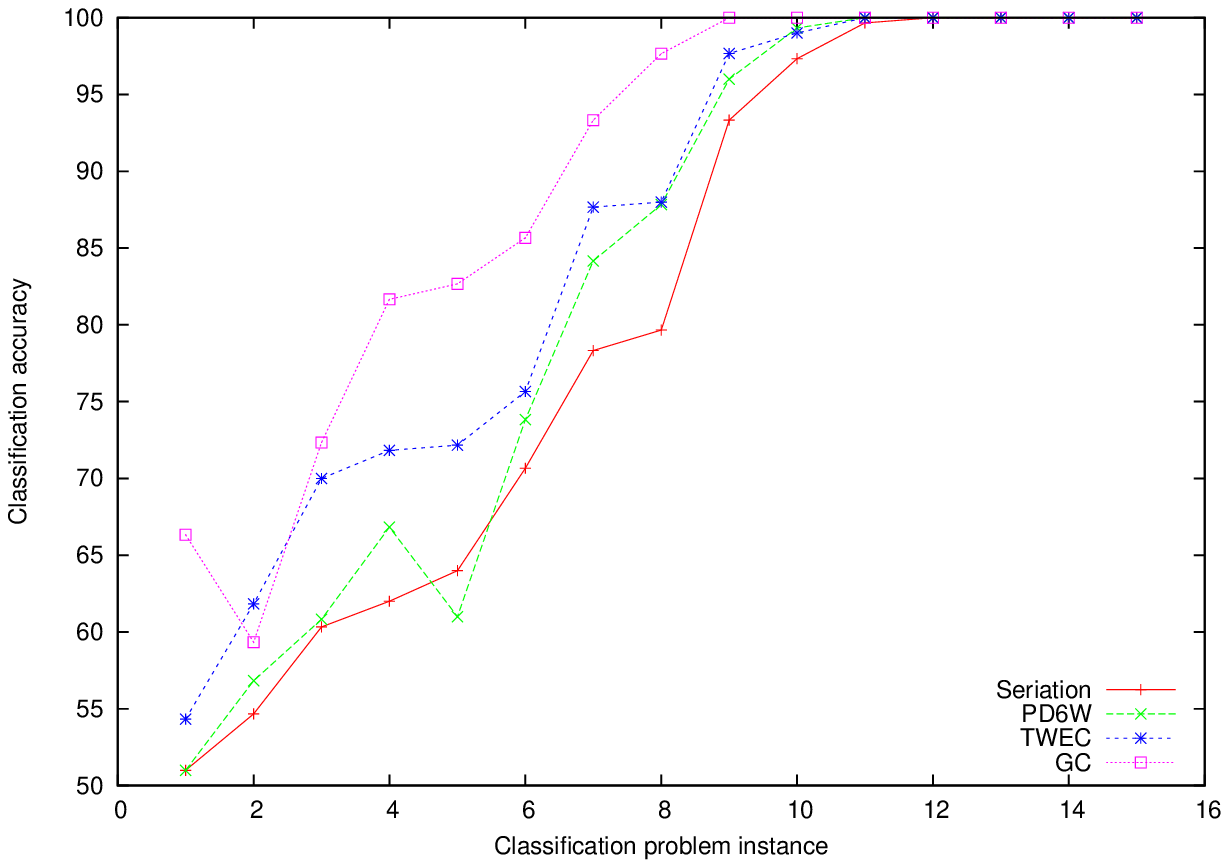}
 \label{fig:graphs_1}
 }
~
\subfigure[Results with $k=3$.]{
    \includegraphics[viewport=0 0 346 243,scale=0.41,keepaspectratio=true]{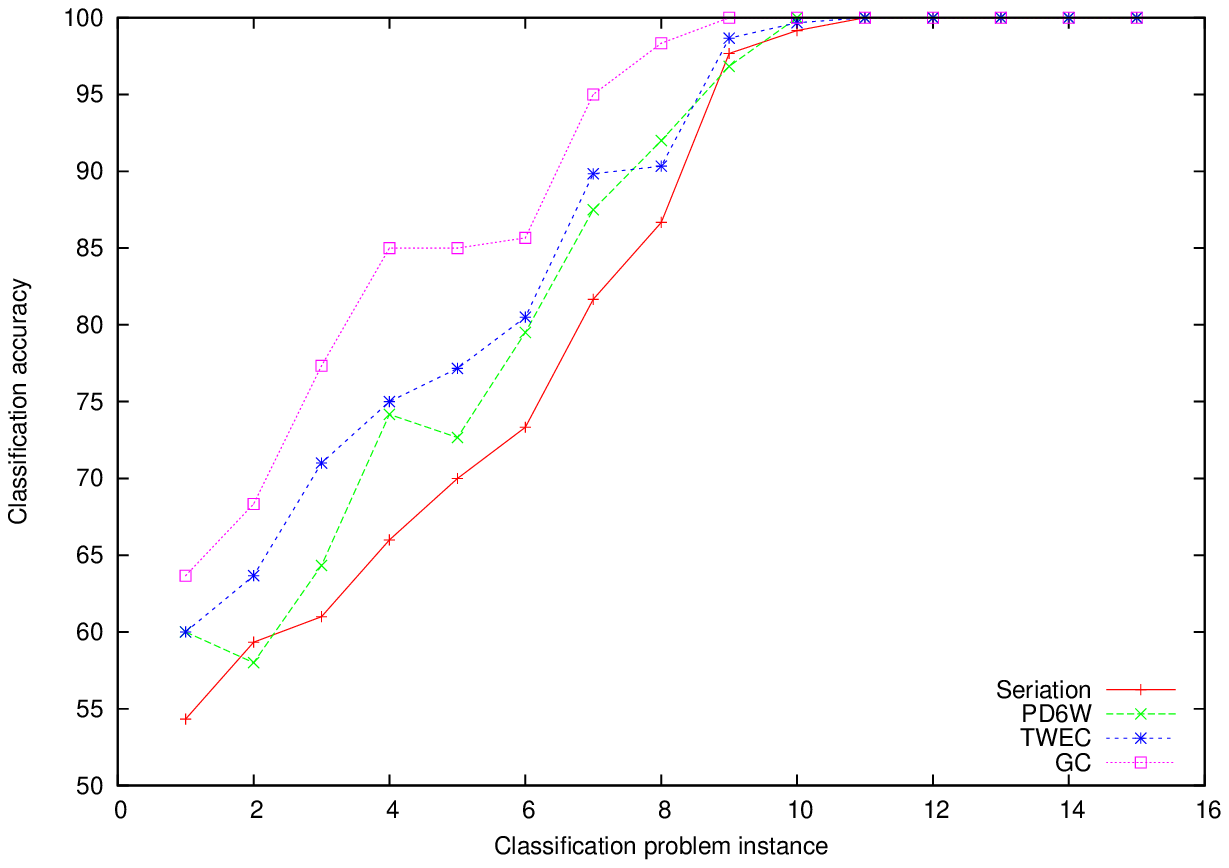}
 \label{fig:graphs_2}
 }
~
\subfigure[Results with $k=5$.]{
    \includegraphics[viewport=0 0 346 243,scale=0.41,keepaspectratio=true]{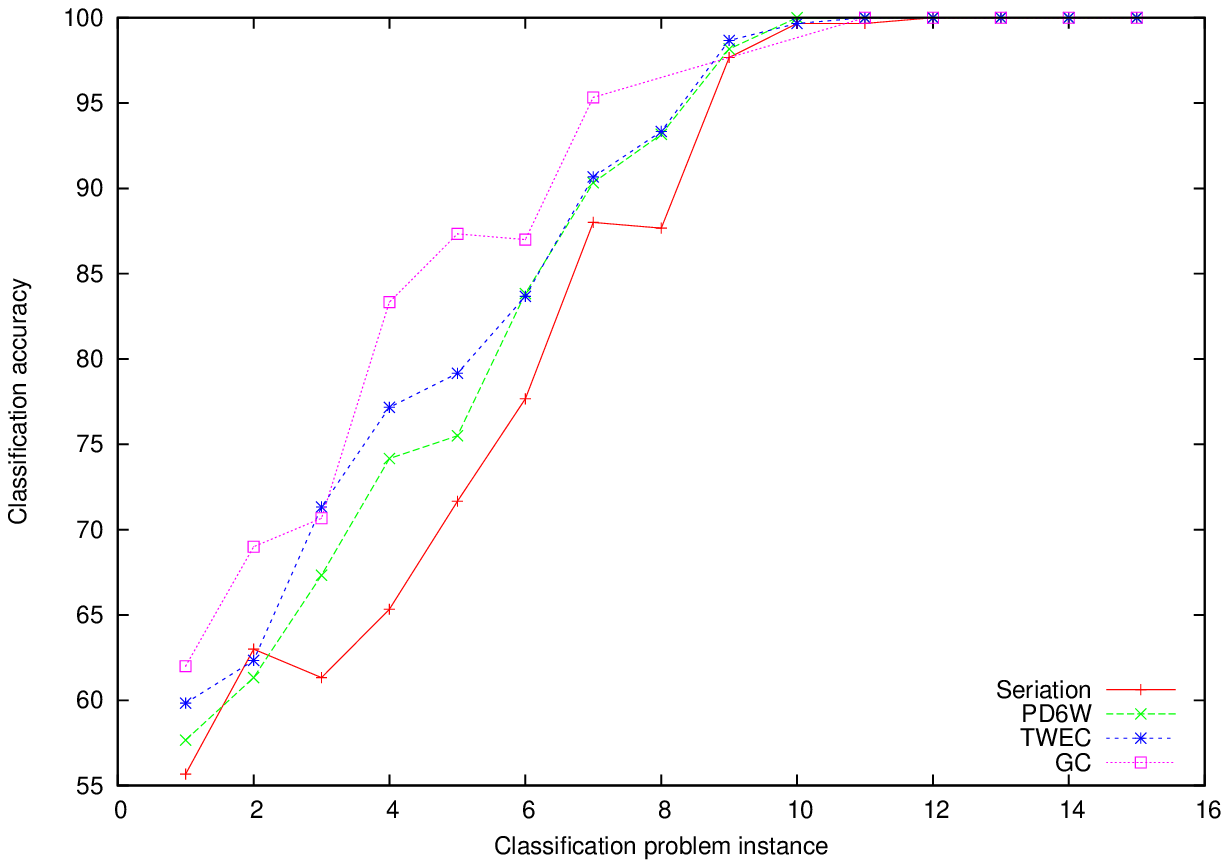}
 \label{fig:graphs_3}
 }

\label{fig:all}
\caption{Test set classification accuracy results of the graph-based classification system over the batch of 15 problem instances.}
\end{figure}

Figures \ref{fig:graphs_1}, \ref{fig:graphs_2}, and \ref{fig:graphs_3} show the obtained performances (test classification accuracy in our case) on the batch of synthetic tests.
Clearly, the system configured with the GC graph matching algorithm achieves the best test set classification accuracy results considering the three values of $k$. TWEC and PD6W obtain comparable results, while the former achieves in general slightly better accuracy values.

\clearpage
\section{Conclusions}
\label{sec:conclusions}

In this paper we have presented the SPARE library as an open source tool to rapidly develop generic pattern recognition applications.
We introduced the underlying design of the library, which is based on the template metaprograming framework offered by the C++ programming language. Main functionalities provided by SPARE have been briefly highlighted. As practical examples of the SPARE features and versatility, we have shown two applications involving clustering and classification problems defined on non-geometric input spaces. The former has been conceived to show the generic MinSOD cluster representation, applied to the context of clustering of real-valued multidimensional sequences.
As a further example, we discussed also the classification of labeled graphs by means of the \textit{k}-NN rule, operating directly on the input space of labeled graphs. All source codes and tutorials can be retrieved together with the latest SPARE release.

Future directions pertain, of course, the development of new algorithms, ranging from new clustering algorithms to additional meta-heuristic global optimization techniques and new graph matching procedures.

\section*{Acknowledgements}

This work was partially supported by the ``POlo per la MObilit\`{a} Sostenibile'' (POMOS), the research center for sustainable mobility, Lazio Region, Italy.

\bibliographystyle{abbrvnat}
\bibliography{/home/lorenzo/University/Research/Publications/Bibliography.bib}
\end{document}